\documentclass[letterpaper, 10 pt, conference]{ieeeconf}  

\IEEEoverridecommandlockouts                              

\usepackage{amsmath,amsfonts}
\usepackage{algorithm}
\usepackage{algpseudocode}
\usepackage{array}
\usepackage[caption=false,font=normalsize,labelfont=sf,textfont=sf]{subfig}
\usepackage{textcomp}
\usepackage{stfloats}
\usepackage{url}
\usepackage{verbatim}
\usepackage{graphicx}
\usepackage{tikz}
\usepackage{multirow}
\usetikzlibrary{angles, quotes}
\usepackage{cite}

\DeclareMathOperator*{\minimize}{minimize}

\title{\LARGE \bf
Optimized and kinematically feasible multi-agent motion planning
}

\author{Anja Hellander$^{1}$, Kristoffer Bergman$^{2}$ and Daniel Axehill$^{1}$
\thanks{*This work was partially supported by the Wallenberg AI, Autonomous Systems and Software Program (WASP) funded by the Knut and Alice Wallenberg Foundation.}
\thanks{$^{1}$A. Hellander and D. Axehill are with the Division of Automatic Control,
        Linköping University, Sweden
        {\tt\small \{anja.hellander, daniel.axehill\}@liu.se}}%
\thanks{$^{2}$K. Bergman is with Saab AB, Link{\"o}ping, Sweden {\tt\small kristoffer.bergman@thernfrst.io}}%
}

\begin{document}

\maketitle
\thispagestyle{empty}
\pagestyle{empty}

\begin{abstract}

Multi-agent motion planning (MAMP) is an important problem for autonomous systems with multiple agents. In this work we propose a two-step method for finding optimized and kinematically feasible solutions to MAMP problems. The first step finds an initial feasible solution using state-of-the-art methods such as conflict-based search (CBS) or priority-based search (PBS), and the second step is an improvement step which improves the solution by solving a multi-phase optimal control problem (OCP) where the initial solution is used to warm-start the solver. We also propose a method for generating motion primitives in an optimized way under the constraint that the primitive durations are all multiples of the same sample time. 

We evaluate our proposed framework on a MAMP problem for tractor-trailer systems. We extend the safe interval path planning with interval projections (SIPP-IP) algorithm so it can handle more general cost functions and larger agents, but our results show that for the tractor-trailer system a simple lattice-based planner performs better due to less conservative collision checks. Our experiments also indicate that CBS performs better than PBS for this system as it achieves a higher success rate in environments with obstacles and had a lower average runtime, although both planners achieve solutions of similar quality after the improvement step.

\end{abstract}

\section{Introduction}
An important problem to solve for autonomous systems is motion planning, i.e., determining what path or trajectory to follow in order to safely and efficiently move the system from a current state to a desired state. An even more challenging problem is multi-agent motion planning (MAMP), where feasible and collision-free trajectories for multiple agents should be planned. In this work, we focus on optimized and kinematically feasible MAMP for agents with complex kinematics such as tractor-trailers.

\subsection{Related Work}

A simpler  version of MAMP is the multi-agent pathfinding (MAPF) problem \cite{stern2019multi, stern2019multi-overview, gao2023review}, where time is discretized and kinodynamic constraints on the agents are not considered. In the most simple form of MAPF, agents move on a grid where each agent occupies one grid cell and at each discretized-step agents will either stay at their current cell or move to a neighbouring cell. More generally, agents move on a graph where vertices represent possible agent states and edges represent possible motions between states.

The optimal MAPF problem, where either the sum or the maximum of the agent times is minimized, is NP-complete \cite{yu2013structure}. The most common approach to optimal MAPF is conflict-based search (CBS) \cite{sharon2015conflict, boyarski2015icbs}, where each agent initially finds a trajectory independently of other agents, and constraints on the trajectories are added until no conflicts between trajectories remain. Other approaches include extending the A* algorithm \cite{hart1968formal} to search in a multi-agent search space \cite{standley2010finding, goldenberg2014enhanced, wagner2015subdimensional}, and the increasing cost tree search (ICTS) that uses a bilevel search \cite{sharon2013increasing, walker2018extended}.

As solving the optimal problem is difficult, in practice suboptimal algorithms are used, such as Prioritized Planning (PP) \cite{erdmann1987multiple, vandenberg2005prioritized, silver2005cooperative} where a path is found for one agent at a time according to a pre-set priority order, and paths of agents with higher priority are seen as dynamic obstacles. An improvement to this is Priority Based Search (PBS) \cite{ma2019searching},  where several partial priority orders are explored. While PP and PBS are often efficient, they are neither optimal nor complete, even if all possible priority orders are explored \cite{ma2019searching}.

There has been an increased interest in extending  the MAPF algorithm so as to be able to solve more complex problems that are more similar to MAMP. Classical MAPF assumes that agents consist of only a single point and therefore occupies only a single grid cell or vertex. A problem for train-like agents is considered in \cite{atzmon2019multi}, where agents cover $k+1$ grid cells: the grid cell currently occupied by the head of the train as well as the last $k$ positions occupied by the head which are now occupied by the train. MAPF for so-called large agents, that have an associated shape and may occupy more than a single grid cell, is considered in \cite{li2019multi}. However, only agent shapes such that the area occupied is the same regardless of orientation such as circles are considered.

MAPF with continuous time and agent motions with non-unit cost that obey some kinematic constraints is considered in \cite{cohen2019optimal}. In addition, the algorithm presented in \cite{andreychuk2022multi} allows for an arbitrary wait time. Both algorithms rely on the Safe Interval Path Planning (SIPP) \cite{phillips2011sipp} algorithm for single-agent motion planning. SIPP, however, makes the assumption that agents can always perform a wait action (i.e., remain at its current grid cell or vertex). In real-life scenarios, agents are typically unable to instantaneously stop or go from standstill to maximum velocity. The trajectories obtained are therefore either not kinodynamically feasible, or they are such that the agent is at standstill at all states corresponding to vertices in the graph in order to allow for wait actions, which leads to suboptimal trajectories. In \cite{ali2023safe} an extension to SIPP called Safe Interval Path Planning with Interval Projection (SIPP-IP) was presented. Unlike SIPP, SIPP-IP only allows wait actions to be performed if the velocity of the agent is zero.

Kinodynamic constraints have also been considered in \cite{wen2022cl, kottinger2022conflict}. In \cite{wen2022cl}, MAMP for car-like vehicles is performed by using CBS in combination with a spatiotemporal Hybrid A* (SHA*). The resulting trajectories are, however, not truly feasible as they assume, e.g., that the vehicles can instantaneously switch between standstill, forward driving at a constant velocity and driving reverse at a constant velocity. A similar approach is taken in \cite{su2025collaborative}, but for small tractor-trailer vehicles and an additional smoothing step is applied to make the trajectories feasible. In \cite{solis2021representation}, CBS is used in combination with probabilistic roadmaps (PRM). Similarly, \cite{kottinger2022conflict} uses CBS, but uses rapidly-exploring random trees (RRT) \cite{lavalle1998rapidly, lavalle2001randomized} as single-agent motion planner and hence the algorithm lacks optimality as well as resolution optimality. 

There are several approaches to optimal motion planning. A common approach to motion planning is to use deterministic or random sampling-based motion planners. Many such algorithms that use random sampling are based on the Rapidly-exploring Random Tree (RRT) algorithm \cite{lavalle1998rapidly} and its asymptotically optimal extension RRT* \cite{karaman2011sampling}. There are extensions of RRT* that can handle dynamical and nonholonomical constraints \cite{karaman2013sampling, stoneman2014embedding}, but the algorithms require the solution of optimal control problems (OCPs) in order to connect states in the generated search tree, which can be computationally demanding for general nonlinear systems \cite{stoneman2014embedding}.

For differentially flat systems, B-splines or Bézier curves are commonly used to smoothen a sequence of previously computed waypoints \cite{yan2024multi}, or as steering functions used within a sampling-based motion planner.

The optimal motion-planning problem can also be formulated as an OCP directly, either as a mixed-integer problem or as a nonlinear problem. In practice, a numerical solver must often be warm-started with a good initial guess to reach convergence. For this reason, one approach is to solve the motion planning in steps where an initial solution is first computed and then used to warm-start the numerical solver. In \cite{li2019tractor} a rough, possibly infeasible, initial trajectory for a tractor-trailer vehicle is computed using Hybrid A* and used as warm-start to solve a sequence of OCPs that gradually add constraints to ensure kinematic feasibility, obstacle avoidance etc. In \cite{bergman2020improved} a lattice-based motion planner \cite{pivtoraiko2009differentially} is used to compute an initial feasible solution which is then used as warm-start to solve an OCP. Similarly, \cite{dahlmann2023local} uses a lattice-based planner to compute an initial path, for which a time-optimal velocity profile is then computed.

Similar optimization approaches have also been applied to MAMP problems. In \cite{toumieh2022decentralized} initial paths for all agents are computed which are then used to compute reference trajectories that are used as input to a mixed-integer quadratic program formulation of model predictive control. This approach, however, is only applicable if the agents have linear (or piecewise affine) dynamics. In \cite{li2024collaborative} reference trajectories that are not guaranteed to be neither dynamically feasible nor collision-free are first computed. These are then smoothed and made to be collision-free by solving an optimization problem within a sequential programming framework.

\subsection{Contributions}

We propose a two-step MAMP approach where in a first step an initial feasible solution is computed using either CBS or PBS, and in a second step a locally optimized solution is computed by posing and solving a multi-phase optimal control problem using the initial feasible solution as initial guess to warm start the numerical solver.

Furthermore, we introduce modifications to SIPP-IP to enable more general cost functions and larger agents. Although SIPP-IP has been shown to outperform a lattice-based A* search as single-agent planner, we show that for complex MAMP problems such as the one for tractor-trailer systems, the lattice-based A* search can perform better.

Finally, we propose a method for generating optimal kinematically feasible motion primitives that are synchronized in time, i.e., have time durations that are a multiple of the same time step length. This is useful for collision checking and is also a requirement for SIPP-IP \cite{ali2023safe}.

\section{Problem Formulation}

Consider $k$ agents operating within the same workspace $W \subset \mathcal{R}^2$. Each agent $i \in \{1, \dots, k\}$ is a nonlinear dynamical system that can be described by
\begin{equation}
\dot{x}_i(t) = f_i(x_i(t), u_i(t)), \quad x(t_0) = x_{i, 0}
\end{equation}
where $x_i \in \mathcal{X}_i \subset \mathcal{R}^{n_i}$ is the state vector, $u_i \in \mathcal{U}_i \subset \mathcal{R}^{m_i}$ is the control input, $t \geq 0$ is the time elapsed, and $x_{i, \text{init}}$ is the initial state of the agent. Let $R_i(x_i) \subset \mathcal{R}^2$ denote the area occupied by agent $i$ when the agent is in state $x_i$. Then, $\mathcal{X}_i$ is the state space for agent $i$ corresponding to $\mathcal{X}_i = \{x_i \in \mathcal{R}^{n_i} | R_i(x_i) \subset W\}$. 

Furthermore, there is an obstacle region $O \in W$ that the agents should not collide with. Thus, each agent $i$ is further constrained as
\begin{equation}
x_i (t) \in \mathcal{X}_{i} \setminus \mathcal{X}_{i, \text{obs}} = \mathcal{X}_{i, \text{free}}
\end{equation} 

\noindent where $\mathcal{X}_{i, \text{obs}}$ denotes the part of the state space that corresponds to the agent intersecting with the obstacle region $O$, i.e., $\mathcal{X}_{i, \text{obs}} = \{x_i \in \mathcal{X}_i | R_i(x) \cap O \neq \emptyset \}$. Additionally, agents must not collide with each  other, i.e., they must obey
\begin{equation}
R_i(x_i(t)) \cap R_j(x_j(t)) = \emptyset \quad  i \neq j
\end{equation}

Given the initial states $x_{i, 0}$ and final states $x_{i, f}$ for each agent, the multi-agent motion-planning problem consists of finding a set of feasible and collision-free trajectories $(x_i(t), u_i(t), t_{i, f})$ such that $x_i(t_0) = x_{i,  0}$ and $x_i(t_{i, f}) = x_{i, f}$ for all agents $i$.

The optimal MAMP problem for $k$ agents can then be posed as

\begin{subequations}
\label{eq:opt-mamp}
\begin{align}
\minimize_{\{x_i(t), u_i(t), t_{i, f}\}_{i=1}^{k}} \quad & \sum_{i=1}^{k} J_i(x_i, u_i, t_{i, f}) \\
\text{s.t.} \quad \dot{x}_i(t) & = f_i(x_i(t), u_i(t)) \\
x_i(t) & \in \mathcal{X}_{i, \text{free}} \quad i = 1, \dots, k \\
u_i(t) & \in \mathcal{U}_i \\
R_i(x_i(t)) \cap R_j(x_j(t)) & = \emptyset \label{subeq:collision} \quad i \neq j \\ 
\label{subeq:init} x_i(t_0) & = x_{i, 0} \\ 
\label{subeq:terminal} x_i(t_{i, f}) & = x_{i, f} 
\end{align}
\end{subequations}

\noindent where the cost functions $J_i$ are defined as
\begin{equation}
\label{eq:costl}
J_i = \int_{t_0}^{t_f} l(x_i(t), u_i(t)) \, dt
\end{equation}
\noindent where $l(x(t), u(t)) \geq 0$ is the user-defined running cost. For example, selecting $l(x, u) = 1$ results in a minimum-time problem.

Finding a feasible and globally optimal solution to \eqref{eq:opt-mamp} is difficult, even for $k=1$, due to the combinatorial aspects of what route to take around obstacles as well as the nonlinearity of the system. For $k > 1$ there is also the combinatorial problem of which agent should be given priority if their paths overlap. For this reason, it is common to use methods that find feasible solutions by solving approximate problems \cite{lavalle2006planning}. One such approach for single-agent motion planning is the lattice-based motion planner \cite{pivtoraiko2009differentially}.

Finding a feasible and locally optimal solution is even more difficult if $k > 1$. If not for constraint \eqref{subeq:collision}, it would be possible to treat \eqref{eq:opt-mamp} as $k$ separate single-agent motion-planning problems. It is common to use methods that find feasible solutions by repeatedly solving (approximate) single-agent motion-planning problems and gradually introduce constraints to ascertain that no collision between agents occurs. Examples of such algorithms include CBS \cite{sharon2015conflict} and PBS \cite{ma2019searching}.

\subsection{The discretized problem}

Lattice-based motion planners \cite{pivtoraiko2009differentially} rely on restricting the control signals to a discrete subset of so-called motion primitives, thereby transforming \eqref{eq:opt-mamp} from a continuous optimization problem to a discrete graph-search problem on a graph $\mathcal{G} = (\mathcal{V}, \mathcal{E})$.

To do so, a discretized state space $\mathcal{X}_d$ is constructed by sampling the state space $\mathcal{X}$ in a regular fashion. Next, the connectivity is selected by deciding which pair of states in $\mathcal{X}_d$ to connect. For each such pair a motion primitive describing a feasible motion (assuming no obstacles) between the states is computed, e.g., by solving an OCP \cite{ljungqvist2019path}. A motion primitive $m$ is defined as
\begin{equation}
m = (x(t), u(t)) \in \mathcal{X} \times \mathcal{U}, t \in [0, T]
\end{equation}
\noindent where $T$ is the time duration of the primitive and $x(0), x(T) \in \mathcal{X}_d$.

The single-agent motion-planning problem can then be approximated as the discrete OCP

\begin{subequations}
\label{eq:single-agent-ocp}
\begin{align}
\minimize_{\{m_i\}_{i=1}^{N}, N} \quad & J_{\text{lat}} =  \sum_{i=1}^{N} L_m(m_i) \\
\text{s.t.} \quad x_1 &= x_0  \\
x_{N+1} &= x_f \\
t_1 &= t_0 \\
x_{i+1} &= \mathbf{f_m}(x_i, m_i) \\
t_{i+1} &= t_i + T_{m_k} \\
m_k & \in \mathcal{P} \\
R(x_i, m_i, t) & \notin \mathcal{X}_{\text{obs}}(t)\quad t \in [t_i, t_{i+1}]
\end{align}
\end{subequations}

\noindent where $\mathbf{f_m}(x, m)$ describes the successor state after motion primitive $\mathbf{m}$ is applied in $\mathbf{x}$.

Successful state-of-the-art multi-agent planners such as CBS and PBS that rely on decomposing the problem into many single-agent planning problems can then solve the approximated problem \eqref{eq:single-agent-ocp}.

\section{The Planning Framework}

\begin{figure}[tb]
\begin{center}
\includegraphics[clip, trim=3cm 4cm 8cm 2cm, width=0.45\textwidth]{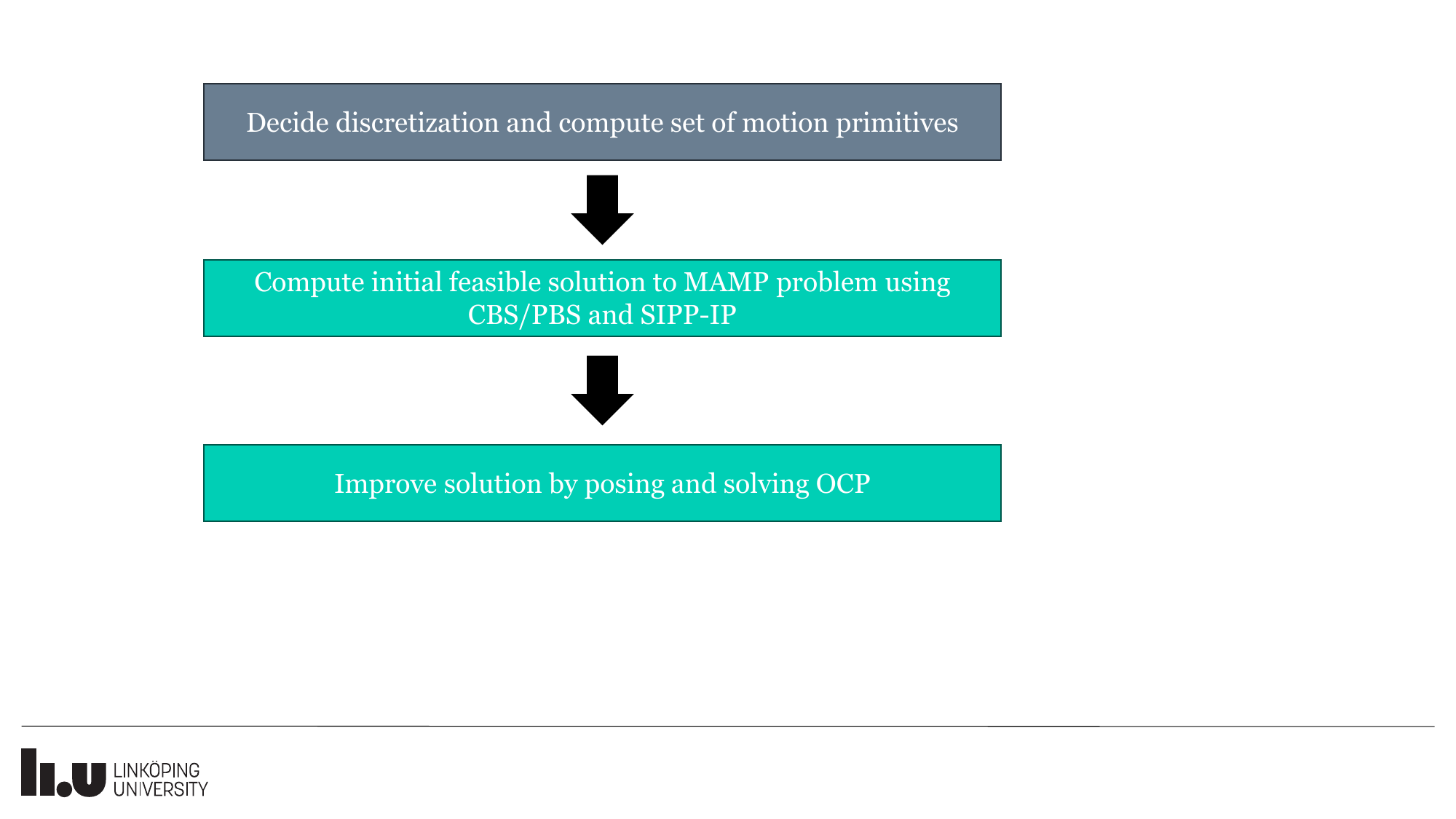} 
\caption{The steps of the proposed framework. Steps in grey are performed off-line and steps in green are performed on-line.} 
\label{fig:flowchart}
\end{center}
\end{figure}

The proposed MAMP framework can be seen as an extension of the single-agent framework proposed in \cite{bergman2020improved}. The underlying principle is to first solve a discretization of the motion-planning problem to obtain an initial feasible solution. As shown in \cite{bergman2020improved}, solving the discretized problem will in general result in a suboptimal solution. For this reason, an improvement step is applied where an OCP is posed and solved using the initial feasible solution found in the previous step as initial guess to warm-start the solver. The steps of the algorithm are shown in Figure \ref{fig:flowchart}.

\section{Computing Time-Synchronous Motion Primitives}\label{sec:primitives}
To enable simple inter-agent collision checking, it is necessary that all motion primitives (represented as a sequence of samples) to share the same discretization, i.e., the time between samples along the motion primitive should be the same for all motion primitives. To achieve this, we propose here a novel extension of the framework in \cite{bergman2019improved}.

Optimal motion primitives are generated by first discretizing the state space, then selecting the connectivity, i.e., deciding which discretized states to connect, and finally solving OCPs to determine a feasible trajectory between the states to connect. As in \cite{bergman2020improved}, we use the same cost function \eqref{eq:costl} for both the primitive generation and the subsequent motion planning.

To compute the primitives, we use an extended version of the framework from \cite{bergman2019improved} where different primitive maneuvers such as parallel or straight are selected and the connectivity is then automatically selected for each maneuver. In the original framework the sample time used by each primitive is a decision variable and is allowed to vary between the primitives. We modify the framework so that we first select a sample time $t_s$ to be used for all primitives. We then select the connectivity and length of the primitives as outlined in Algorithm \ref{alg:primitive}.

For each pair of initial state and primitive maneuver a set of candidate terminal states are generated. For each candidate terminal state a motion primitive is generated by solving the corresponding OCP without any constraint on the sample time (line 4). Based on the time duration $T$ of the motion primitive two candidate primitive durations are computed as $t_s \left\lfloor \frac{T}{t_s} \right\rfloor$ and $t_s\left\lceil \frac{T}{t_s} \right\rceil$ respectively. For each candidate duration the OCP is solved again with an additional constraint on the time duration and the sample time (line 7), resulting in a candidate motion primitive. The candidate primitive with the best cost function value is then selected.

\begin{algorithm}[tb]
\caption{Primitive generation}
\label{alg:primitive}
\begin{algorithmic}[1]
\State Input: initial state $x_0$, primitive maneuver $q$, sample time $t_s$
\State minCost = $\infty$, primitive = $\emptyset$
\For{$x_T \in $ candidateTerminalStates($x_0, q$)}
	\State $p$ = optimizePrimitive($x_0, x_T$)
	\State $T$ = duration($p$)
	\For{$N \in \{ \left\lfloor \frac{T}{t_s} \right\rfloor, t_s\left\lceil \frac{T}{t_s} \right\rceil\}$}
		\State $\bar{p} = $ optimizePrimitive($x_0, x_T, N, t_s$)
		\If{cost($\bar{p}$) $<$ minCost}
			\State minCost = cost($\bar{p}$)
			\State primitive = $\bar{p}$
		\EndIf
	\EndFor
\EndFor
\State \Return primitive
\end{algorithmic}
\end{algorithm}

\section{Computing an Initial Feasible Solution}

This section describes the framework for computing an initial feasible solution. 

\subsection{Multi-agent planner}

As  multi-agent motion planner we consider both CBS and PBS. CBS and PBS are both based on searching a graph where nodes contain a set of constraints on the trajectories of each agent as well kinematically feasible trajectories that obey these constraints. A general  outline is shown in Algorithm \ref{alg:mamp}. First, the root node is created without any constraints and a plan consisting of trajectories for each agent is computed (Line 3) by ignoring constraint (7b) in (7). Next, nodes are iteratively removed from the open list (Line 6) and checked for conflicts, i.e., collisions between agents. A conflict between a pair of agents $a_i, a_j$ is selected (Line 10) and two new nodes are created to which constraints on the trajectories of $a_i$ and $a_j$, respectively, are added so as to solve the conflict (Line 13). The trajectories of the agents are updated so that the constraints are obeyed (Line 14) and the new nodes are inserted into the open list.

In CBS, each constraint that is added applies only at a given point in time. Depending on implementation, constraints may specify that an agent is not allowed to be in a certain state at a certain time, or is not allowed to apply a certain motion primitive from a certain state at a certain time.

In PBS, constraints are in the form of partial priority orders that specify which agents take priority. Agents with lower priority treat agents with higher priority as dynamic obstacles and plan trajectories that avoid them.

\begin{algorithm}[tb]
\caption{General MAMP algorithm}
\label{alg:mamp}
\begin{algorithmic}[1]
\Procedure{MAMP}{}

\State Root.constraints = $\emptyset$
\State Root.plan = updatePlan(Root, Root.constraints)
\State $Q$.insert(Root)
\While{not $Q$.empty()}
	\State $n = Q$.pop()
	\If{$n$.conflicts() $= 0$}
		\State \Return $n$.plan
	\EndIf
	\State $C = n$.firstConflict()
	\For{agent $i$ involved in $C$}
		\State $n' = n$
		\State $n'\text{.constraints} = n.\text{constraints} \cup \text{makeConstraint}(C, i)$
		\State $n'$.plan = updatePlan($n'$, $n'$.constraints)
		\State $Q$.insert($n'$)
	\EndFor
\EndWhile

\EndProcedure
\end{algorithmic}
\end{algorithm}

\subsection{Single-agent motion planner}

As single-agent motion planner we propose to use a lattice-based motion planner where the state has been augmented with time.

We also extended SIPP-IP to allow for general cost functions $J_i$ on the form \eqref{eq:costl}. In the original SIPP-IP, each search node corresponds to an agent state and an associated time interval $[t_l, t_u]$ and the cost-to-come of a search node is $g(n) = t_l$. To perform the desired extension, we assume that each motion primitive is associated with a cost and that the cost-to-come of a node is $g(n) = t_l + \bar{g}(n)$, where $\bar{g}(n)$ is the sum of the costs of the motion primitives necessary to reach $n$. The proposed extension introduces more book-keeping to check if a node has been seen before with lower cost or not. In the original algorithm it is sufficient to store the time intervals that have been explored for each state. If a new time interval $[t_l, t_u]$ is explored where another time interval $[t_l, t]$ has already been explored, only the remaining interval $[t, t_u]$ will be considered. With the proposed modifications, however, it is possible to reach a state at a time that overlaps with an already explored time interval, but with a lower cost. For each explored time interval it is then necessary to keep track of the lowest cost-to-come seen for that interval. Only repeated time intervals with an equal or higher cost are then discarded.

SIPP-IP handles collisions by dividing the environment into a grid cell. Each motion primitive stores a list of what cells are swept during what intervals, and collision checking is performed by checking against the unsafe intervals of each grid cell. The original SIPP-IP implementation assumed that the agent occupied only a single grid cell at the end of a motion primitive. This is not realistic for large agents such as tractor-trailers. We therefore introduce the possibility to have larger agents by allowing multiple cells to be terminal cells. With only a single terminal cell it is straightforward to determine $t_u$ for a state with zero velocity as it will be the upper bound of the safe interval for the terminal cell. When multiple terminal cells are allowed, as in this work, $t_u$ is instead selected as the lowest such upper bound.

There are two major differences between the two single-agent motion-planning algorithms considered.
First, a search node in SIPP-IP corresponds to an agent state and a time interval whereas a search node in the lattice-based planner corresponds to an agent state and a single point in time. This suggests that SIPP-IP is more efficient as fewer states are needed. The second difference is collision checking. SIPP-IP uses grid cells and intervals as described above. The lattice-based planner instead checks for collisions at each time step by approximating the agents and static obstacles with covering discs and performing circle-to-circle collision checks. These circles are used when computing the cells that are swept by the motion primitives for SIPP-IP as well. As a cell is considered occupied as long as any part of the circle is inside it this implies that SIPP-IP has a more conservative collision avoidance, which in general will result in plans with lower performance. In particular, two agents with circles sweeping through the same cell will be considered to collide even though the circles may not actually overlap.

\section{Improvement using Optimal Control}

In this section we describe the improvement step, where we propose to use numerical  optimal control to improve the initial feasible solution, in line with what has been proposed for single-agent motion-planning problems in \cite{bergman2020improved}. 

Initially this might seem straightforward, but posing \eqref{eq:opt-mamp} as on overarching OCP that can be solved by off-the-shelf solvers turns out to be non-trivial. In the single-agent case, practical implementations typically rely on introducing variables $x_n, n=0,  1, \dots, N$ where $x_n = x(nT_s), T_s=\frac{t_f}{N}$ and the terminal time $t_f$ is a decision variable in the optimization problem. This representation makes it easy to enforce the initial and terminal constraints \eqref{subeq:init} and \eqref{subeq:terminal} by constraining the values of $x_0$ and $x_N$, respectively. It is not immediately obvious how to extend this to a multi-agent setting where the terminal times $t_{i, f}$ are not necessarily the same. If on the one hand the trajectories $x_i(t)$ are discretized independently of each other and $t_{i, f}$ are introduced as decision variables it is not straightforward to encode the collision-avoidance constraint \eqref{subeq:collision} as it is not straightforward to determine which discretized variables that correspond to the same instant in absolute time as the $t_{i, f}$ change values during the optimization. On the other hand, if all trajectories use the same $T_s$, then it is straightforward to implement \eqref{subeq:collision}, but difficult to implement \eqref{subeq:terminal}. That is, if $x_{i, N_i} = x_{i, f}$ is enforced, then there is little freedom in reducing $t_{i, f}$ as reducing the terminal time for one agent must result in the terminal times for all other agents being reduced in equal proportion. The alternative is to find some other way to encode \eqref{subeq:terminal}. 

\begin{figure}[tb]
\begin{center}
\includegraphics[clip, trim=5.0cm 5.0cm 5.0cm 3.0cm, width=0.45\textwidth]{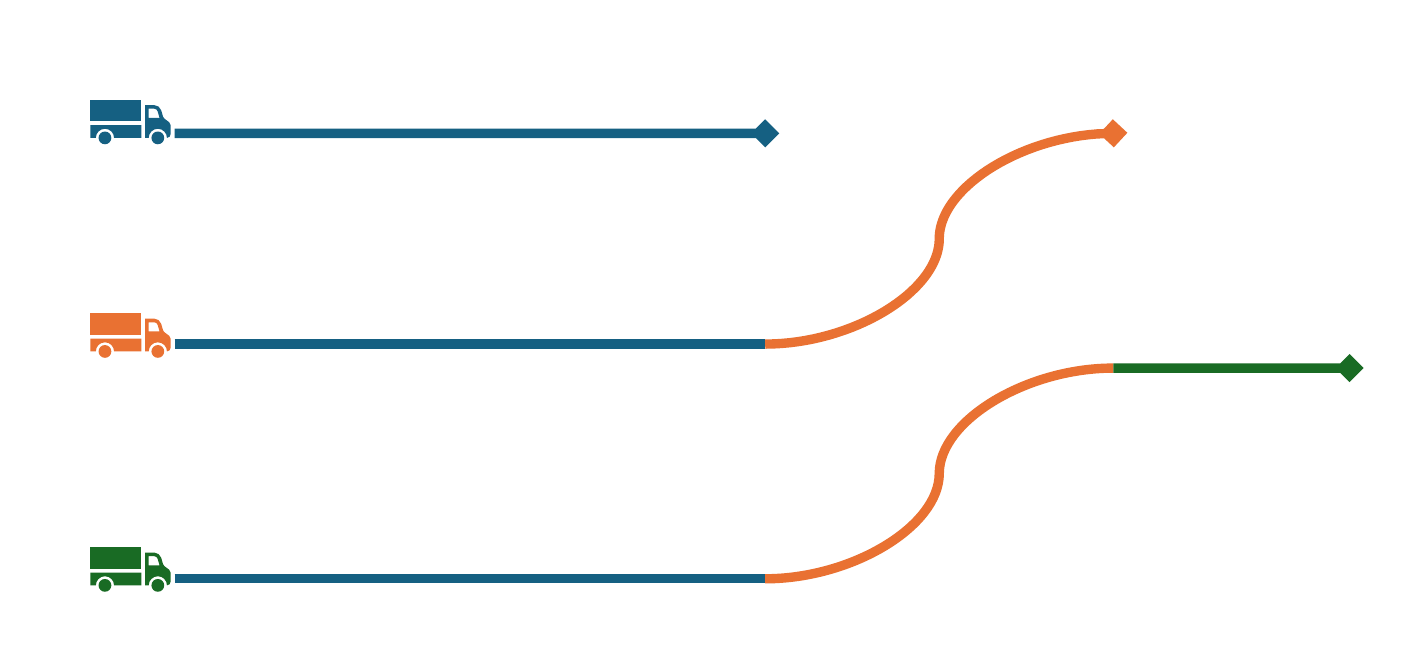} 
\caption{Illustration of phases for three agents. Each color represents a different phase. Each phase covers the time until one of the agents reaches its goal. The initial positions are marked by the truck symbols, and the final positions are indicated by the arrowheads.} 
\label{fig:faser}
\end{center}
\end{figure}

The solution we propose in this work is to pose the problem  \eqref{eq:opt-mamp} as a multi-phase OCP. Using the trajectories $(x_i(t), u_i(t), t_{i, f})$ obtained in the first step we assume without loss of generality that the agents are ordered such that $t_{1, f} \leq t_{2, f} \leq \dots \leq t_{k, f}$, i.e., that if $i < j$, then agent $i$ reaches its goal no later than agent $j$. If not, we simply reorder them. We divide the problem into $M=k$ phases where the first phase covers the time until agent $i=1$ reaches its goal and each subsequent phase $m$ covers the time between the time when agent $m-1$ reaches its goal and agent $m$ reaches its goal. An illustration of this is shown in Figure \ref{fig:faser}. For each phase $m$ we introduce the decision variable $t_m$ denoting the time duration of the phase. For each phase it is then possible to use the same time discretization for all agent trajectories to allow for easy collision checking, while different phases having different discretization allows for some agent trajectories to decrease in time while others increase as the time duration of each phase is independent of the others.

Note that the order in which the agents reach their goal will be the same as in the initial feasible solution from the first step. Similarly, the combinatorial aspect of how to navigate around static obstacles as well as other agents is also implicitly encoded by the initial feasible solution and will be maintained.

The OCP to solve can then be formulated as

\begin{subequations}
\begin{align}
\minimize_{\{x_i(t), u_i(t), t_{i}\}_{i=1}^{k}} \quad & \sum_{m=1}^{k} J_m(x, u, t_m) \\
\text{s.t.} \quad x_{0, i}(t_0) &= x_{i, 0} \\
x_{m, m}(t_m) &= x_{m, f} \\
x_{m+1, i}(t_m) &= x_{m, i}(t_m), i \geq m+1 \\
\dot{x}_{m, i}(t) & = f_i(x_i(t), u_i(t)), \quad t \in [t_{m-1}, t_m] \\
R_i(x_{m, i}(t)) \cap R_j(x_{m, j}(t)) &= \emptyset \quad i \neq j
\end{align}
\end{subequations}

\noindent where $J_m$ is the cost function defined in \eqref{eq:costl}, (8c) decodes the constraint that for each phase $m$ the $m$th agent reaches its goal and constraint (8d) ensures continuity between phases.

%

\section{Numerical Results}

In this section the proposed planning algorithms are implemented and evaluated in problems with truck and trailer systems in unstructured environments. The single-agent motion planner, CBS, and PBS planners are implemented in C++. The single-agent motion planner uses synchronized motion primitives that have been optimized as described in Section \ref{sec:primitives}. It uses A* search guided by a heuristic function in the form of a heuristic lookup table (HLUT) \cite{knepper2006high} that has been precomputed offline. The HLUT contains the optimal costs of trajectories in obstacle-free space from all initial states $x_0 \in \mathcal{X}_d$ positioned at the origin to all final states $x_f \in \mathcal{X}_d$ positioned within a square centered around the origin with side length $\rho$. Note that the truck and trailer system is position and rotation invariant.

The computation of the motion primitives and the improvement step are both implemented in Python using CasADi \cite{andersson2019casadi} with IPOPT \cite{wachter2006implementation} and the ma57 solver from HSL.

\subsection{Vehicle models}

\begin{figure}[tb]
\centering

\begin{tikzpicture}
\coordinate (origo) at (0,0);
\draw (origo) -- node[above] {$L_3$} ++  (25:3cm) coordinate (A) -- node[left] {$L_2$} ++(70:1.5cm) coordinate (B) -- node[below] {$M_1$} ++ (25:0.5cm) coordinate (C) -- node[above] {$L_1$} ++ (25:2cm) coordinate (D);

\draw [dashed] (origo) -- (0:1.5cm) coordinate (th3);
\draw [dashed] (C) -- ++ (0:1.5cm) coordinate (th1);
\draw [dashed] (B) -- ++ (70:1.5cm) coordinate (b2);
\draw [dashed] (A) -- ++ (250:1.5cm) coordinate (b3);
\draw [dashed] (D) -- ++ (25:1.5cm) coordinate (E);
\draw [dashed] (D) -- ++ (55:1.5cm) coordinate (F);

\draw [very thick] (D) -- ++ (55:0.15cm) coordinate (G);
\draw [very thick] (D) -- ++ (235:0.15cm) coordinate (H);

\draw (origo) -- (115:0.4cm) coordinate (o1);
\draw (origo) -- (295:0.4cm) coordinate (o2);
\draw (A) -- ++ (160:0.4cm) coordinate (a1);
\draw (A) -- ++ (340:0.4cm) coordinate (a2);
\draw (C) -- ++ (115:0.4cm) coordinate (c1);
\draw (C) -- ++ (295:0.4cm) coordinate (c2);
\draw [thick] (o1) -- ++ (25:0.07cm);
\draw [thick] (o1) -- ++ (205:0.07cm);
\draw [thick] (o2) -- ++ (25:0.07cm);
\draw [thick] (o2) -- ++ (205:0.07cm);
\draw [thick] (a1) -- ++ (70:0.07cm);
\draw [thick] (a1) -- ++ (250:0.07cm);
\draw [thick] (a2) -- ++ (70:0.07cm);
\draw [thick] (a2) -- ++ (250:0.07cm);
\draw [thick] (c1) -- ++ (25:0.07cm);
\draw [thick] (c1) -- ++ (205:0.07cm);
\draw [thick] (c2) -- ++ (25:0.07cm);
\draw [thick] (c2) -- ++ (205:0.07cm);

\pic [draw, ->, "$\theta_3$", angle eccentricity=2.0] {angle = th3--origo--A};
\pic [draw, ->, "$\theta_1$", angle eccentricity=2.0] {angle = th1--C--D};
\pic [draw, <-, "$\beta_2$", angle eccentricity=1.3, angle radius=1cm] {angle = C--B--b2};
\pic [draw, ->, "$\beta_3$", angle eccentricity=2.0] {angle = origo--A--b3};
\pic [draw, ->, "$\alpha$", angle eccentricity=2.0] {angle = E--D--F};

\end{tikzpicture}

\caption{An illustration of the truck and trailer system.}
\label{fig:schematic}
\end{figure}
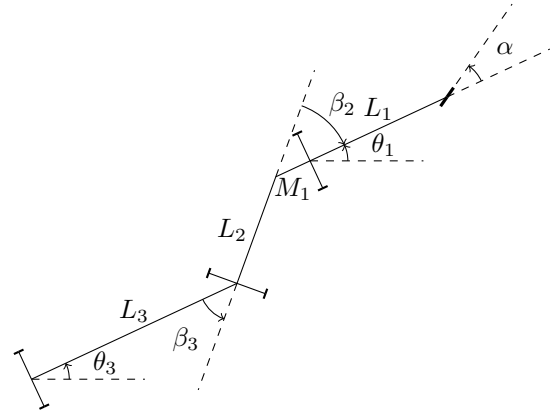

Each truck and trailer system is a general 2-trailer with a car-like truck and off-axle hitching \cite{altafini2002hybrid, ljungqvist2019path}. Each such system has three vehicle components: a car-like tractor, a dolly, and a semitrailer as illustrated in Figure \ref{fig:schematic}. The state vector of the system is

\begin{equation}
\begin{aligned}
\mathbf{x} & = 
\begin{bmatrix}
\mathbf{z}^T & \alpha & \omega & v_1 & a_1 \\ 
\end{bmatrix}^T \\
\mathbf{z} & = 
\begin{bmatrix}
x_3 & y_3 & \theta_3 & \beta_3 & \beta_2 \\
\end{bmatrix}^T
\end{aligned}
\end{equation}

\noindent where $\mathbf{z}$ describes the state of the truck and trailer, $(x_3, y_3)$ is the position of the semitrailer, $\theta_3$ is the orientation of the semitrailer, $\beta_3$ is the joint angle between the semitrailer and dolly and $\beta_2$ is the joint  angle between the dolly and the truck. As in \cite{bergman2020optimization} the state vector has been augmented with $\alpha$ and $\omega$, the steering angle and steering angle rate, respectively, of the truck as well as the longitudinal velocity $v_1$ and acceleration $a_1$ of the truck.

With the control signal vector
\begin{equation}
\mathbf{u} =
\begin{bmatrix}
u_\omega & u_a
\end{bmatrix}^T
\end{equation}

\noindent the system can be modelled as \cite{bergman2020optimization}:

\begin{equation}
\begin{aligned}
\dot{\mathbf{x}} & = 
\begin{bmatrix}
v_1 f(\mathbf{z}, \alpha) & \omega & u_\omega & a_1 & u_a \\
\end{bmatrix}^T
\end{aligned}
\end{equation}

\noindent where
\begin{equation}
f(\mathbf{z}, \alpha) =
\begin{bmatrix}
\dot{x}_3 & \dot{y}_3 & \dot{\theta}_3 & \dot{\beta}_3 & \dot{\beta}_2 \\
\end{bmatrix}^T
\end{equation}
\begin{equation}
\begin{aligned}
 \dot{x}_3 & = \cos{\beta_3} (1 + \frac{M_1}{L_1} \tan{\beta_2} \tan{\alpha}) \cos{\theta_3} \\
 \dot{y}_3 & = \cos{\beta_3} (1 + \frac{M_1}{L_1} \tan{\beta_2} \tan{\alpha}) \sin{\theta_3} \\
 \dot{\theta}_3 & = \frac{\sin{\beta_3}\cos{\beta_2}}{L_3} (1 + \frac{M_1}{L_1} \tan{\beta_2} \tan{\alpha}) \\
 \dot{\beta}_3 & = \cos{\beta_2}(\frac{1}{L_2}( \tan{\beta_2} - \frac{M_1}{L_1} \tan{\alpha} ) - \\
 & \quad \quad \frac{\sin{\beta_3}}{L_3} (1 + \frac{M_1}{L_1} \tan{\beta_2} \tan{\alpha}) ) \\
\dot{\beta}_2 &=  \frac{\tan{\alpha}}{L_1} -  \frac{\sin{\beta_2}}{L_2}+ \frac{M_1}{L_1 L_2} \cos{\beta_2} \tan{\alpha} \\
 \end{aligned}
\end{equation}

The cost function, which is used both in the improvement step and during the search for the initial feasible solution, is selected as
\begin{equation}
l(\mathbf{x}, \mathbf{u}) = 1 + \frac{1}{2}(\alpha^2 + 10 \omega^2 + a_1^2 + \mathbf{u}^T \mathbf{u}).
\end{equation}

\noindent 

All agents have the same geometry as described in \cite{ljungqvist2019path}.

\subsection{Simulation Results}

To evaluate the performance of the proposed framework we first randomly generate 100 problem instances for $n=2, 3, 4, 5$ agents in an obstacle-free $200 \times 200$ map. All problem instances are generated so that there is no overlap between initial positions or between terminal positions, and so that there exists a feasible trajectory for each agent if all other agents are ignored. Secondly, we randomly generate 100 additional problem instances in a $200 \times 200$ map where static obstacles are randomly generated as well. The HLUT radius is $\rho = 200$. We implemented both CBS and PBS with both SIPP-IP and a lattice planner as single-agent planner using the same set of optimized motion primitives. Each planner was given up to 100$\,s$ to solve each problem instance.

We first compared SIPP-IP and the lattice-based planner by applying them to the generated problems as if they were single-agent problems. In total there were 500 such single-agent problems for the obstacle-free map and 500 such problems for the obstacle maps. The average computation times and plan costs are shown in Table \ref{tab:sipp-vs-lattice}. It can be seen that the lattice-based planner is much faster than SIPP-IP when there are no obstacles present, but slower when obstacles are present. It can also be seen that both planners return nearly the same average cost without obstacles (the cost for SIPP-IP is slightly higher), but when obstacles are present the average cost is 14$\,\%$ higher for SIPP-IP. This higher cost can be explained by the difference in collision avoidance between the planners. As the collision avoidance of SIPP-IP is more conservative, plans that are feasible for the lattice-based planner may be discarded as infeasible by SIPP-IP which is then forced to select a plan with higher cost instead. Note that the obstacle-free map is not truly obstacle-free as the boundaries of the map serve as obstacles, which explains the difference in cost for that map as well.

The reason that SIPP-IP is faster in the presence of obstacles is likely due to the efficient representation of time as intervals. SIPP-IP is therefore able to conclude for a whole time interval that applying a certain primitive from a certain state is impossible due to collision. The lattice-based planner on the other hand is only able to conclude that the primitive can not be applied at a given point in time and may therefore try to increasingly apply a wait primitive and then try the original primitive again.


\begin{table}
\caption{Comparison of the average performance of the proposed single-agent planners.}
\label{tab:sipp-vs-lattice}
\centering
\begin{tabular}{|c | c | c | c | c|} 
 \hline
 \multirow{2}{*}{Planner} & \multicolumn{2}{c|}{Obstacle-free} & \multicolumn{2}{c|}{Obstacles} \\
 & Cost & Time [s] & Cost & Time [s] \\
 \hline
 Extended SIPP-IP & 131.30 & 0.12 & 144.66 & 2.05\\
 \hline
 Lattice-based & 130.80 & 0.008 & 126.19 & 3.54\\
 \hline
\end{tabular}
\end{table}

\begin{figure}[tb]
\begin{center}
\includegraphics[clip, trim=3.8cm 8.5cm 4cm 9cm, width=0.4\textwidth]{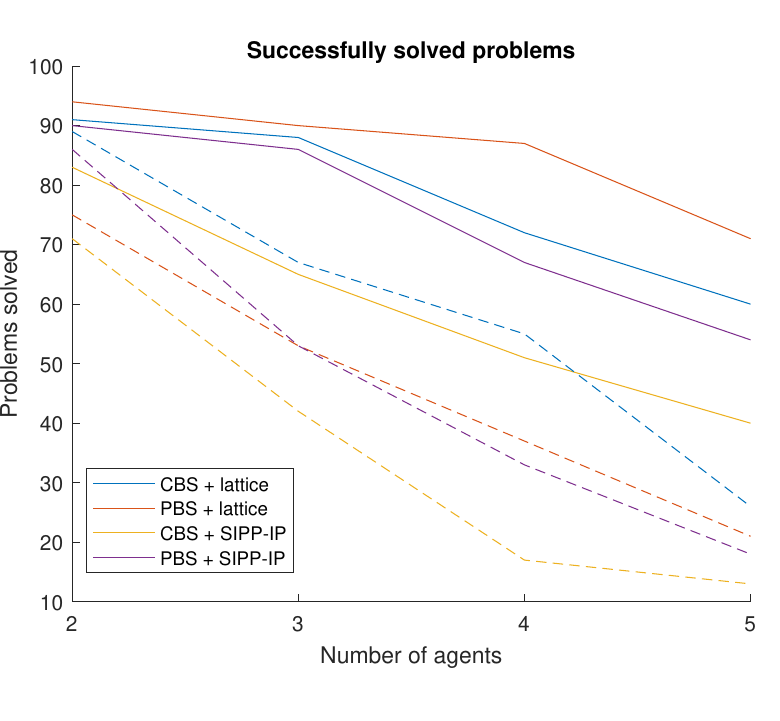} 
\caption{The number  of problems successfully solved by the first step within the allowed time limit. Fully drawn lines show the results for the obstacle-free map and dashed lines show the results for the maps with obstacles.} 
\label{fig:sipp-vs-lattice}
\end{center}
\end{figure}

Next, we applied the first step of the proposed multi-agent framework to the generated problems. Figure \ref{fig:sipp-vs-lattice} shows the number of problem instances that were successfully solved by the planners within the allotted time of 100$\,$s. It can be seen that for a given multi-agent planner the lattice-based planner always performed better than SIPP-IP. The difference in success rate can be explained by SIPP-IP having a more conservative collision avoidance, so that solutions that are feasible for the lattice-based planner are discarded as infeasible by SIPP-IP. This is the case both for static obstacles as well as dynamic obstacles arising from the trajectories of other agents. This can cause SIPP-IP to either fail to find a feasible solution, or to be so slow that it fails to find a feasible solution within the time limit. Similarly, we see that for a given single-agent planner PBS usually performs better than CBS, with the exception being in the case with obstacles for the lattice-based planner.

In Table \ref{tab:sipp-vs-lattice-pbs} a comparison for SIPP-IP and the lattice-based planners considering the problems that both planners managed to solve with PBS is shown. The lattice-based planner is faster than SIPP-IP in the obstacle-free case as well as when there are obstacles present. It also always performs better in terms of cost function. In the obstacle-free case the difference is only a few percent, but with obstacles present it is around $15\,\%$.

\begin{table}
\caption{Comparison of the average performance of the SIPP-IP and the lattice-based planner as single-agent planners for PBS considering the problems both planners managed to solve.}
\label{tab:sipp-vs-lattice-pbs}
\centering
\begin{tabular}{|c| c | c | c | c | c|} 
 \hline
 \multirow{2}{*}{Planner} & \multirow{2}{*}{Agents} & \multicolumn{2}{c|}{Obstacle-free} & \multicolumn{2}{c|}{Obstacles} \\
 & & Cost & T [s] & Cost & T [s] \\
 \hline
 SIPP-IP & 2 & 265.71& 2.10 & 288.58 & 7.95\\
 \hline
 Lattice-based & 2 & 263.87 & 0.25& 257.47 & 4.16\\
 \hline
 SIPP-IP & 3 & 398.15 & 7.58 & 423.96 & 13.23\\
 \hline
 Lattice-based & 3 & 393.16 & 1.48 & 376.27& 5.98\\
 \hline
 SIPP-IP & 4 & 521.31 & 5.31 & 579.38 & 37.50 \\
 \hline
 Lattice-based & 4 & 513.41 & 4.61& 520.09& 11.95 \\
 \hline
 SIPP-IP & 5 & 641.06 & 10.62 & 678.71 & 26.42\\
 \hline
 Lattice-based & 5 & 626.86 & 7.29 & 602.48 & 16.52 \\
 \hline
\end{tabular}
\end{table}

For the evaluation of the proposed two-step MAMP approach we choose to use the lattice-based planner as it had a higher success rate. We evaluated both PBS and CBS as multi-agent planners. For all problems where a feasible solution was successfully found, we applied the continuous improvement step where the numerical solver was given 100$\,$s to improve the solution.  

\begin{figure}[tb]
\begin{center}
\includegraphics[clip, trim=3.8cm 8.5cm 4cm 9cm, width=0.4\textwidth]{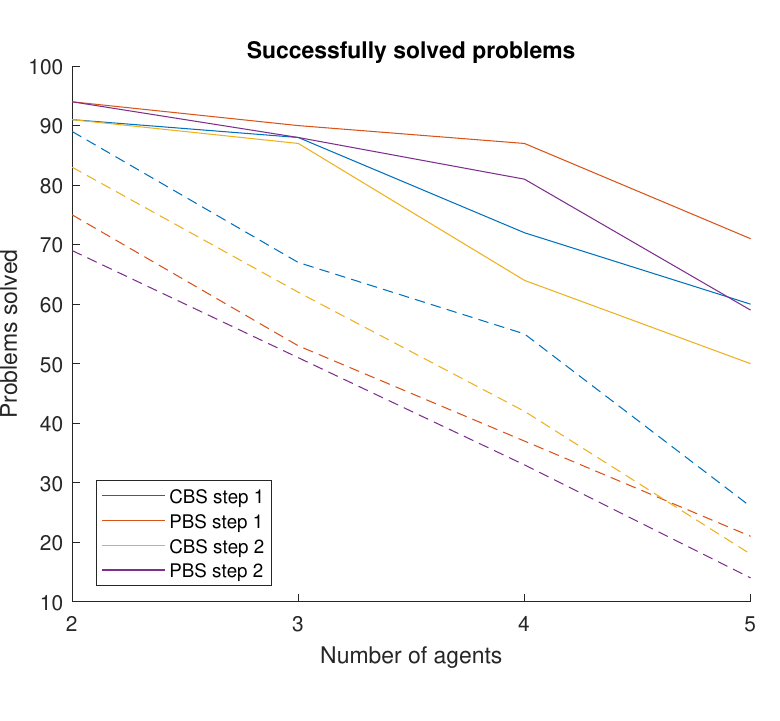} 
\caption{The number  of problems successfully solved by the two planners. Successfully solved during step 2 means that the numerical optimization solver was able to improve the solution. Fully drawn lines show the results for the obstacle-free map and dashed lines show the results for the maps with obstacles.} 
\label{fig:cbs-vs-pbs}
\end{center}
\end{figure}

\begin{figure}[tb]
\begin{center}
\includegraphics[clip, trim=3.8cm 8.5cm 4cm 9cm, width=0.40\textwidth]{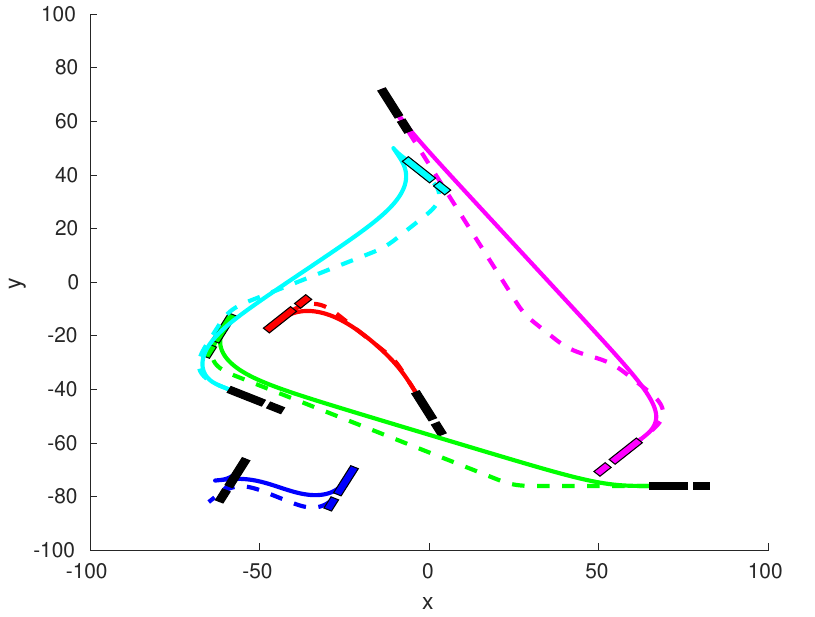} 
\caption{Example of the resulting plans for an example with four agents. The black tractor-trailers show the initial positions and final positions are colored. Dashed lines show the trajectories after the first step and fully drawn lines show the final solution after the improvement step. It is clear that artefacts from discretization in the initial solution have been removed in the final solution.} 
\label{fig:example}
\end{center}
\end{figure}

Figure \ref{fig:cbs-vs-pbs} shows the number of problems that were successfully solved by the planners. For the obstacle-free case, it can be seen that PBS solves a higher number of problems than CBS does and that the gap increases as the number of agents increases. It can also be seen that for $n=2$ agents the optimization step is able to improve all initial solutions, but as the number of agents increases the number of problems where an improved solution is found (within the time limit) decreases. The number of problems where a feasible solution is found during the first step also decreases, so the optimization step still succeeds for the majority of the problems where it is applied. Table \ref{tab:tab1_both} shows the results for the problems solved by both planners, to facilitate a direct comparison on execution time and cost function values. As seen in Table \ref{tab:tab1_both}, for the first step CBS is faster than PBS for all number of agents. As CBS is (resolution) optimal and PBS is not, it returns solutions with lower cost, but the difference in cost function value is very small. After the improvement step, however, the average cost is sometimes slightly lower for PBS and sometimes slightly lower for CBS. An example of resulting trajectories after the first and second steps is shown in Figure \ref{fig:example}.

\begin{table}
\caption{Comparison of CBS and PBS in an obstacle-free environment. Problems solved by both planners. Computation time is denoted with $t$, cost function value with $J$ and the number of problems solved within the time limit with $r$. Subscripts $1$ and $2$ denote the first and second steps, respectively.}
\label{tab:tab1_both}
\centering
\begin{tabular}{|c c c c c c c c|} 
 \hline
 N & Planner & $t_1$ & $t_2$ & $J_1$ & $J_2$ & $r_1$ & $r_2$\\ [0.5ex] 
 \hline \hline
 2 & CBS & 0.08 & 2.92 & 262.79 & 221.45 & 86 & 86\\ 
 \hline
 2 & PBS & 0.56 & 3.05 & 264.01 & 221.09 & 86 & 86\\
 \hline
 3 & CBS & 1.03 & 13.06 & 393.43 & 335.36 & 82 & 81\\
 \hline
 3 & PBS & 1.29 & 13.09 & 393.45 & 335.89 & 82 & 80\\
 \hline
 4 & CBS & 1.07 & 24.49 & 501.52 & 433.10 &  71 & 63\\
 \hline
 4 & PBS & 2.11 & 26.32 & 501.74 & 426.81 & 71 & 68\\
 \hline
 5 & CBS & 3.49 & 39.92 & 635.44 & 552.54 & 49 & 42\\
 \hline
 5 & PBS & 7.40 & 41.09 &  635.86 & 558.41 & 49 & 40\\
 \hline
\end{tabular}
\end{table}

As seen in Figure \ref{fig:cbs-vs-pbs}, both planners solve less problems in the case with obstacles than for the obstacle-free case. Unlike the obstacle-free case, CBS performs better than PBS for all number of agents. As in the obstacle-free case the number of problems where the solution is improved (within the time limit) during the improvement step decreases as the number of agents increases. A comparison between CBS and PBS is presented in Table \ref{tab:tab2_both} where only the problems solved by both planners in the first step are considered. The time needed to find a solution has increased. CBS is still faster than PBS. As in the obstacle-free case, CBS finds solutions with a slightly lower average cost after the first step but after the improvement step PBS sometimes achieves solutions with slightly lower average cost. This is likely due to randomness rather than any inherent property of PBS, but it does demonstrate that a lower cost function value after the first step does not necessarily translate to a lower cost function value after the second step.

\begin{table}
\caption{Comparison of CBS and PBS in an environment with obstacles for the problems solved by both planners. Computation time is denoted with $t$, cost function value with $J$ and the number of problems solved within the time limit with $r$. Subscripts $1$ and $2$ denote the first and second steps, respectively.}
\label{tab:tab2_both}
\centering
\begin{tabular}{|c c c c c c c c|} 
 \hline
 $N$ & Planner & $t_1$ & $t_22$ & $J_1$ & $J_2$ & $r_1$ & $r_2$\\ [0.5ex] 
 \hline \hline
 2 & CBS & 1.68 & 3.26 & 251.59 & 211.21 & 75 & 69\\ 
 \hline
 2 & PBS & 4.21 & 3.27 & 251.67 & 210.68 & 75 & 69\\
 \hline
 3 & CBS & 1.57 & 14.01 & 372.52 & 314.22 & 49 & 47\\
 \hline
 3 & PBS & 6.48 & 13.61 & 372.54 & 314.27 & 49 & 47\\
 \hline
 4 & CBS & 3.98 & 23.38 & 499.45 & 422.38 & 32 & 28\\
 \hline
 4 & PBS & 6.13 & 22.36 & 499.67 & 421.03 & 32 & 28\\
 \hline
 5 & CBS & 11.77 & 45.25 & 583.41 & 497.76 & 13 & 10\\
 \hline
 5 & PBS & 16.86 & 39.36 & 583.94 & 506.69 & 13 & 9\\
 \hline
\end{tabular}
\end{table}
\section{Conclusions and Future Work}

In this work we consider a multi-agent motion planning (MAMP) problem for tractor-trailers. The problem is challenging due to the complex kinematics of the vehicles as well as due to the vehicles being comprised of several connected bodies. We propose a framework for optimized MAMP consisting of two steps. In the first step an initial, feasible and collision-free, solution is computed using conflict-based search (CBS) or priority-based search (PBS). The second step is an improvement step where a multi-phase optimal control problem (OCP) is posed and the solution found in the previous step is used to warm-start the solver.

We have also proposed a framework for automatically generating optimal motion primitives with the same time discretization, a requirement for the single-agent planner safe interval path planning with interval projection (SIPP-IP), in an optimized way. The cost function used to generate the primitives is consistently also the cost function used by the continuous improvement step.

We have implemented and evaluated the proposed framework on a large number of randomly generated problems for tractor-trailer systems. The results show that using a lattice-based planner as single-agent planner performs better than SIPP-IP, likely due to a less conservative collision avoidance. The results also show that although lacking completeness guarantees, PBS is able to solve a higher number of problems than CBS in an obstacle-free environment and can achieve solutions of similar quality after the improvement step. However, the results also show that CBS is faster than PBS on the problems they both solve and solves a higher number of problems in environments with obstacles.

Future work includes improving the computational efficiency of the implementation of the improvement step by using distributed optimization and parallelization. Another possible line of research is the addition of traffic rules that the agents should obey.

\bibliographystyle{ieeetr}
\bibliography{refs_IV}

\end{document}